\documentclass[lettersize,journal]{IEEEtran}
\usepackage{amsmath,amsfonts}
\usepackage{algorithmic}
\usepackage{algorithm}
\usepackage{array}
\usepackage{xcolor}
\usepackage{romannum}
\usepackage[caption=false,font=normalsize,labelfont=sf,textfont=sf]{subfig}
\usepackage{textcomp}
\usepackage{stfloats}
\usepackage{url}
\usepackage{verbatim}
\usepackage{graphicx}
\usepackage{cite}
\usepackage{enumerate}
\usepackage[colorlinks, linkcolor=blue, anchorcolor=blue, citecolor=blue]{hyperref}
\usepackage{lipsum}
\usepackage{enumitem}
\usepackage{amsmath}
\usepackage{multirow}
\usepackage{tabularx}
\usepackage{bm} 
 \usepackage{booktabs}
 \usepackage{threeparttable} 
\begin{document}
\pagenumbering{arabic}
\title{Theoretical Data-Driven MobilePosenet: Lightweight Neural Network for Accurate Calibration-Free 5-DOF Magnet Localization}

\author{
Wenxuan Xie$^{1}$, Yuelin Zhang$^{1}$, Jiwei Shan$^{1}$, Hongzhe Sun$^{1}$, Jiewen Tan$^{1}$, Shing Shin Cheng$^{1,*}$\\
\thanks{Research reported in this work was supported in part by Research Grants Council (RGC) of Hong Kong (CUHK 14217822 and CUHK 14207823), Multi-Scale Medical Robotics Center, InnoHK, Innovation and Technology Commission of Hong Kong  (ITS/235/22), and The Chinese University of Hong Kong Direct Grant. The content is solely the responsibility of the authors and does not reflect the views of the sponsors.}
\thanks{$^{1}$Department of Mechanical and Automation Engineering and T Stone Robotics Institute, The Chinese University of Hong Kong, Hong Kong.}%
\thanks{$^{*}$Corresponding author: Shing Shin Cheng (sscheng@cuhk.edu.hk).}%
}

\markboth{Journal of \LaTeX\ Class Files,~Vol.~14, No.~8, August~2021}%
{Shell \MakeLowercase{\textit{et al.}}: A Sample Article Using IEEEtran.cls for IEEE Journals}


\maketitle

\begin{abstract}

Permanent magnet tracking using the external sensor array is crucial for the accurate localization of wireless capsule endoscope robots. Traditional tracking algorithms, based on the magnetic dipole model and Levenberg-Marquardt (LM) algorithm, face challenges related to computational delays and the need for initial position estimation. More recently proposed neural network-based approaches often require extensive hardware calibration and real-world data collection, which are time-consuming and labor-intensive. To address these challenges, we propose MobilePosenet, a lightweight neural network architecture that leverages depthwise separable convolutions to minimize computational cost and a channel attention mechanism to enhance localization accuracy. Besides, the inputs to the network integrate the sensors' coordinate information and random noise, compensating for the discrepancies between the theoretical model and the actual magnetic fields and thus allowing MobilePosenet to be trained entirely on theoretical data. Experimental evaluations conducted in a \(90 \times 90 \times 80\) mm workspace demonstrate that MobilePosenet exhibits excellent 5-DOF localization accuracy ($1.54 \pm 1.03$ mm and $2.24 \pm 1.84^{\circ}$) and inference speed (0.9 ms) against state-of-the-art methods trained on real-world data. Since network training relies solely on theoretical data, MobilePosenet can eliminate the hardware calibration and real-world data collection process, improving the generalizability of this permanent magnet localization method and the potential for rapid adoption in different clinical settings.

\end{abstract}


\section{Introduction}
\label{sec:introduction}
\IEEEPARstart{W}{ireless} capsule endoscopy (WCE) has become a widely adopted non-invasive tool for gastrointestinal examination \cite{weitschies1997high,than2012review}. However, the autonomous movement of the capsule inside the body presents significant challenges for real-time tracking of its position and orientation. To address this, permanent magnet tracking has been proposed, which involves embedding a permanent magnet into the capsule and employing an external magnetic sensor array to locate it. This approach enables localization of the capsule within the body, enhancing the safety and efficiency of the treatment process.

Modeling the magnetic field using numerical integration \cite{su2023amagposenet, song20146} or the magnetic dipole model \cite{8809206, hu2010cubic} are two common approaches in permanent magnet tracking based on the sensor array. Numerical integration methods used to solve for six degree-of-freedom (DOF) poses are computationally intensive and non-convex \cite{su2023amagposenet}, and are thus not suitable for real-time tracking. In contrast, the magnetic dipole model simplifies the magnetic field calculation, enabling real-time tracking and retrieval of 5-DOF pose information \cite{8809206,9170792,9631373,hu2010cubic}.

When combined with sensor measurements, the modeled magnetic field data can be used to estimate the pose of the permanent magnet. Among the various optimization-based algorithms proposed to solve this inverse problem for pose estimation, the Levenberg-Marquardt (LM) algorithm is widely used due to its robust performance \cite{8809206,hu2010cubic,su2017investigation,xu2020novel}. However, as a local optimization method, LM requires an initial position guess for iterative optimization. Deviations between the initial guess and the true position can degrade both the efficiency and accuracy of LM. To mitigate this issue, several methods, such as linear algorithms \cite{hu2010cubic}, particle swarm optimization (PSO) \cite{song20146}, and hybrid feedforward neural networks (HFFNNs) \cite{qin2022hffnn}, have been proposed to improve the initial guess. While PSO and HFFNN offer efficient initial guesses for the LM algorithm, they are computationally intensive. For instance, PSO can take up to 830 seconds to estimate the pose of a 6-DOF ring magnet \cite{song20146}, and HFFNNs require extensive computational resources and training time.
 
The computational complexity of traditional methods has prompted recent research to investigate neural-network-based methods for addressing the challenges of permanent magnet tracking \cite{su2023amagposenet, sebkhi2019deep, lv2019pkbpnn, guo2022improved, 10599224}. Sebkhi et al. \cite{sebkhi2019deep} developed a fully automated 5D positional stage, generating a dataset containing 1.7 million samples. However, the data collection process is labor-intensive and time-consuming, requiring almost six months for a dataset with 1 mm positional resolution. To alleviate the need for large datasets, Yao et al. \cite{guo2022improved} utilized a residual neural network (ResNet) to process triaxial magnetometer data, interpreting the three axes as analogous to RGB channels, i.e., treating each axis as a separate color channel. They collected 5,312 data points spanning six magnet poses for training and evaluation. Ren et al. \cite{su2023amagposenet} proposed generating a synthetic dataset based on prior knowledge for model pre-training, followed by fine-tuning with a limited set of real-world samples. Nonetheless, fine-tuning with sparse real data introduces the risk of overfitting. 

It is noteworthy that the localization accuracy of both the aforementioned LM-based and neural network-based methods is highly dependent on the quality of sensor data. Sensor noise can significantly degrade tracking accuracy, typically caused by manufacturing errors in the sensors and magnets. Several studies \cite{su2017investigation, li2009new, hu2006calibration} have proposed methods to calibrate the noise resulting from these errors. However, these approaches are often complex and unsuitable for large-scale clinical applications. Therefore, the development of magnetic tracking methods that can tolerate manufacturing errors without requiring calibration is necessary.

To address challenges such as dependence on initial guesses, overfitting risks, and the requirement for real-world datasets and hardware calibration, we propose MobilePosenet, a lightweight neural network designed for real-time permanent magnet tracking. MobilePosenet leverages depthwise separable convolutions and a channel attention mechanism, significantly reducing computational costs while maintaining the ability to capture essential features. By incorporating the additional sensor location and artificially added random noise as part of the newly proposed input representation, MobilePosenet effectively mitigates discrepancies between the idealized magnetic dipole model and real-world magnetic fields, enabling the network to be trained exclusively on theoretical data generated from the magnetic dipole model with highly accurate localization result. Comprehensive experiments demonstrate that MobilePosenet achieves superior 5-DOF localization performance compared with  state-of-the-art (SOTA) models trained on real-world data. The main contributions of this work are summarized as follows:

\begin{enumerate}[label=\arabic*)]
 
  \item MobilePosenet is proposed as a lightweight architecture for 5-DOF permanent magnet localization, integrating a channel attention mechanism and depthwise separable convolutions. The channel attention mechanism enhances accuracy by assigning higher weights to features most relevant for positioning through a reweighting process, while depthwise separable convolutions reduce computational cost by decomposing standard convolutions into depthwise and pointwise convolutions. These combined techniques result in improvement in positioning accuracy and execution speed compared to existing methods.
  
  \item A novel training method that utilizes only the theoretical dataset generated by the magnetic dipole model is proposed for the first time. The network input combines theoretical magnetic flux densities with the coordinate information of triaxial magnetometers, with random noise added to compensate for discrepancies between the magnetic dipole model and the real magnetic field. This input methodology, integrated with a channel attention mechasnism, eliminates the need for sensor array calibration and real-world data collection, enabling MobilePosenet to be easily adopted in different clinical settings.

\end{enumerate}

\section{METHODS}
\label{sec:method}
\subsection{Magnetic Dipole Model}
\label{subsec:MDM}
In this work, an external sensor array is employed to locate a cylindrical permanent magnet. Previous research by Hu \cite{Hu2006} demonstrates that when the distance between the permanent magnet and the sensor exceeds eight times the magnet's radius, the magnetic dipole model can approximately simulate its magnetic field, thus enabling its 5-DOF pose estimation.

\begin{figure}[t]
\centering
\includegraphics[width=2.5in]{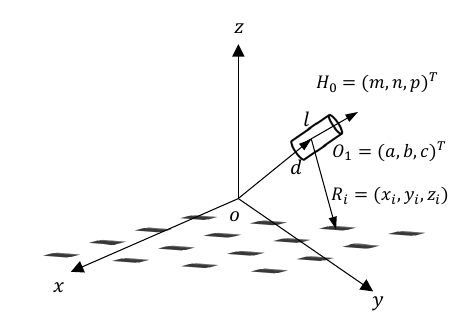}
\caption{Coordinate system for magnet’s localization: ${{O}_{1}}$ is the central position of the magnet, ${{H}_{0}}$ is the direction of the magnet. ${{R}_{i}}$ is the distance between the $i-th$ sensor and ${{O}_{1}}$, $l$ is magnet length and $d$ is magnet diameter.}
\label{fig_4}
\vspace{-6 mm}
\end{figure}

Fig. \ref{fig_4} depicts a cylindrical permanent magnet in space with length $l$, diameter $d$, and surface magnetization $M$. Let ${{X}_{i}} = {{({{x}_{i}} - a, {{y}_{i}} - b, {{z}_{i}} - c)}^{T}}$ denote the vector representing the relative position of a spatial point \((x_i, y_i, z_i)\) with respect to the magnet's center at \((a, b, c)\). The magnetic flux density at this point can be calculated as:
\begin{equation}
\label{deqn_ex1a7}
{{B}_{i}}={{B}_{T}}(\frac{3({{H}_{0}}\cdot {{X}_{i}}){{X}_{i}}}{{{R}_{i}}^{5}}-\frac{{{H}_{0}}}{{{R}_{i}}^{3}})
\end{equation}
where ${{B}_{T}}=({{\mu }_{r}}{{\mu }_{0}}\pi {{r}^{2}}lM)/4\pi $; 
${{\mu }_{r}}$ is the relative permeability of the medium; ${{\mu }_{0}}$ is the magnetic permeability of air; r is the radius of the magnet; ${{H}_{0}}$ is the normalized vector representing the orientation of the magnet’s magnetism.

Suppose there are $N$ sensors in the space and the $i-th$ sensor is located at $({{x}_{i}},{{y}_{i}},{{z}_{i}})$, the magnetic flux density at the $i-th$ sensor location can be represented by:
\begin{equation}
\label{deqn_ex1a6}
{{B}_{i}}={{B}_{ix}}i+{{B}_{iy}}j+{{B}_{iz}}k
\end{equation}
where ${{B}_{ix}}$, ${{B}_{iy}}$, and ${{B}_{iz}}$ represent the three components of the magnetic induction intensity of the point. Specifically, it is expressed as follows:
\begin{equation}
\label{deqn_ex1a5}
\scriptsize
{{B}_{ix}}={{B}_{T}}\left\{ \frac{3\left[ m\left( {{x}_{i}}-a \right)n\left( {{y}_{i}}-b \right)+p\left( {{z}_{i}}-a \right) \right]\left( {{x}_{i}}-a \right)}{R_{i}^{5}}-\frac{m}{R_{i}^{3}} \right\}
\end{equation}
\begin{equation}
\label{deqn_ex1a4}
\scriptsize
{{B}_{iy}}={{B}_{T}}\left\{ \frac{3\left[ m\left( {{x}_{i}}-a \right)n\left( {{y}_{i}}-b \right)+p\left( {{z}_{i}}-a \right) \right]\left( {{y}_{i}}-b \right)}{R_{i}^{5}}-\frac{n}{R_{i}^{3}} \right\}
\end{equation}
\begin{equation}
\label{deqn_ex1a3}
\scriptsize
{{B}_{iz}}={{B}_{T}}\left\{ \frac{3\left[ m\left( {{x}_{i}}-a \right)n\left( {{y}_{i}}-b \right)+p\left( {{z}_{i}}-a \right) \right]\left( {{z}_{i}}-c \right)}{R_{i}^{5}}-\frac{p}{R_{i}^{3}} \right\}
\end{equation}
where ${{R}_{i}}=\sqrt{{{\left( {{x}_{i}}-a \right)}^{2}}+{{\left( {{y}_{i}}-b \right)}^{2}}+{{\left( {{z}_{i}}-c \right)}^{2}}}$.

It is worth noting that the direction vector ${{H}_{0}}$ does not correspond directly to the rotation angle. Suppose that the yaw angle is $\varphi$, the pitch angle is $\theta$, the roll angle is $\phi$, then ${{H}_{0}}$ can be calculated as follows:
\begin{equation}
\label{deqn_ex1a1}
{{[m,n,p]}^{T}}={{R}_{B}}\cdot {{[0,0,1]}^{T}}
\end{equation}
where ${{R}_{B}}$ is the rotation matrix defined as follows:
\begin{equation}
\label{deqn_ex1a2}
{{R}_{B}}=Rot(z,\phi )Rot(y,\theta )Rot(x,\varphi )
\end{equation}

\begin{figure*}[t]
\centering
\includegraphics[width=\textwidth,keepaspectratio]{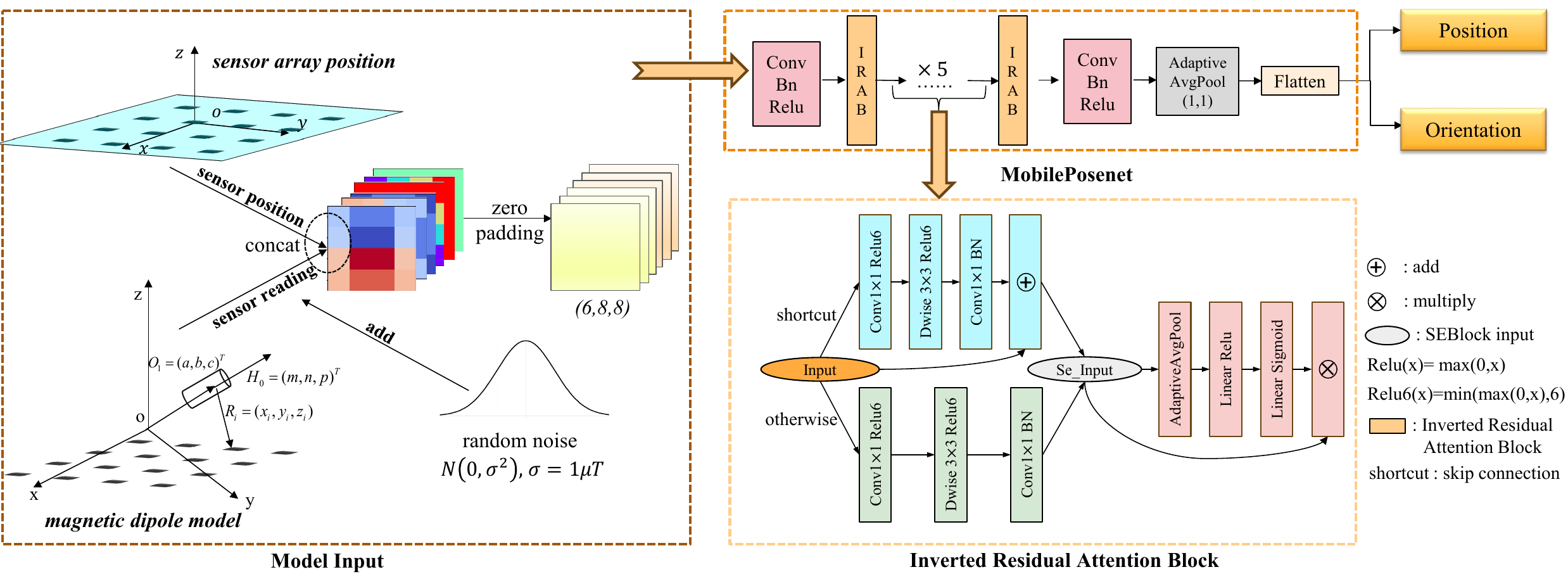}
\caption{Structure and Workflow of MobilePosenet: The network performs end-to-end predictions of the magnet's pose. The input consists of sensor readings and coordinates, while the output consists of the magnet's position and orientation. In the network architecture, ConvBnRelu represents a composite operation that includes a Conv2d layer, Batch Normalization, and the ReLU activation function. IRAB denotes an Inverted Residual Attention Block. AdaptiveAvgPool refers to an adaptive average pooling layer. Dwise indicates depthwise separable convolution. SEBlock (Squeeze-and-Excitation Block) is a channel attention mechanism that enhances feature extraction by reweighting the input channels.}
\label{fig_sim}
\vspace{-6 mm}
\end{figure*}

\subsection{Theoretical Data Training and Neural Network Input Design}
\label{subsec:TDTNNID}
In this work, each sensor in the sensor array is a triaxial magnetometer. The measurements of the triaxial magnetometer can be considered as a three-channel (RGB) image pixel. Previous studies such as \cite{su2023amagposenet,guo2022improved} utilize convolutional neural networks to capture correlations among adjacent sensors in the sensor array to predict the magnet's pose. However, these studies rely solely on sensor array readings as input for the neural networks, typically requiring real-world datasets to train or fine-tune the models. As noted in \cite{sebkhi2019deep}, collecting real-world data for neural network training is a time-consuming task, making it difficult to sample these data at small enough intervals for a dense dataset.

To address these challenges, our proposed MobilePosenet is trained exclusively using theoretical datasets generated from the magnetic dipole model. This approach avoids the challenges of training or fine-tuning the network using real data. 
However, in practical localization scenarios, the magnetic dipole model demonstrates inherent limitations in accurately simulating the magnetic fields generated by permanent magnets, primarily due to factors such as imperfections in sensor manufacturing, environmental noise, and non-uniform magnetization. Training neural networks solely on theoretical datasets risk overfitting, which may degrade localization performance. Therefore, to provide the neural network with more localization-relevant information and thereby improve accuracy, we draw inspiration from the use of sensor array positions in the LM algorithm. Our neural network input comprises not only the theoretical magnetic flux densities of the magnetometers but also their corresponding positions, organized into two $3 \times 4 \times 4$ tensors:

1: Each magnetometer is considered as a three-channel pixel point, with the channels representing \(({{B}_{x}}, {{B}_{y}}, {{B}_{z}})\).

2: The position of each magnetometer is considered as a three-channel pixel point, with the channels representing the \((X, Y, Z)\) coordinates of the sensor.

Given that 16 magnetometers are used to form the sensor array, each component of the input is resized to \(3 \times 4 \times 4\). During model training, the two input components are concatenated to form a final input of size \(6 \times 4 \times 4\), where the six channels are denoted by \(({{B}_{x}}, {{B}_{y}}, {{B}_{z}}, X, Y, Z)\). Besides, random noise is added to the three magnetic flux density channels, following a normal distribution \(\mathcal{N}(0, \sigma^2)\), where \(\sigma\) is the standard deviation and \(\sigma = 1\,\mu T\).

The proposed input methodology enables the neural network to capture critical information, including sensor coordinates and random noise, thereby enabling the network to recognize the discrepancies between the theoretical model and real magnetic fields. This approach mitigates the loss of information due to the absence of real-world data in the training or fine-tuning process, thereby contributing to enhanced positioning accuracy.

\subsection{Inverted Residual Attention Block}
\label{subsec:IRABSUB}
To enhance the localization accuracy of the neural network while minimizing computational cost, depthwise separable convolution is used as the base layer of MobilePosenet while the channel attention mechanism is introduced to recalibrate the extracted features. As shown in Fig.~\ref{fig_sim}, we leverage the inverted residual block introduced in \cite{sandler2018mobilenetv2} and integrate it with the squeeze-and-excitation block (SEBlock) \cite{hu2018squeeze} to construct the Inverted Residual Attention Block (IRAB). This lightweight module forms the foundational block of MobilePosenet. The information flow is shown in Table \ref{tab:IRAB}.  

In the IRAB, the inverted residual block based on depthwise separable convolution significantly reduces computational overhead, while the SEBlock enhances model accuracy. Given an input tensor $L_i$ with dimensions $h_i \times w_i \times d_i$, where $h_i$ and $w_i$ denote the spatial resolution and $d_i$ represents the number of input channels, a convolutional kernel of size $k \times k$ is applied to produce an output tensor $L_j$ with dimensions $h_i \times w_i \times d_j$, where $d_j$ is the number of output channels. The computational cost of this operation using standard convolution is $h_i \times w_i \times d_i \times d_j \times k^2$, while depthwise separable convolution reduces this cost to $h_i \times w_i \times d_i \times (k^2 + d_j)$ by splitting the standard convolution into deep convolution and point convolution, making the computational expense in IRAB approximately $\frac{1}{k^2}$ of that incurred by standard convolutions. While the reduction in computational complexity is primarily driven by the inverted residual block, the SEBlock plays a crucial role in improving localization accuracy by adaptively reweighting feature channels. This reweighting mechanism enhances key features relevant to permanent magnet localization while suppressing irrelevant or redundant information. It allows the network to flexibly select important features in response to variations in the magnetic field and the position of the permanent magnet, thereby improving both robustness and generalization. Furthermore, when addressing the noise present in the data, the adaptive weighting mechanism enables the IRAB to prioritize higher-quality data, assigning greater importance to these parts of the input during the localization process, further enhancing localization accuracy.

\begin{table}[t]
\caption{Information Flow of Inverted Residual Attention Block}
\vspace{-4pt}
\centering
\label{tab:IRAB}
\begin{tabularx}{\columnwidth}{lll}
    \toprule
\textbf{Input}                 & \textbf{Operator}             & \textbf{Output}               \\ \midrule
$k \times h \times w$ & 1x1 conv2d, ReLU6   & $tk \times h \times w$ \\ 
$tk \times h \times w$ & 3x3 dwise s=s, ReLU6 & $tk \times (h/s) \times (w/s)$ \\ 
$tk \times (h/s) \times (w/s)$ & linear 1x1 conv2d & $k' \times (h/s) \times (w/s)$ \\ 
$k' \times (h/s) \times (w/s)$ & SEBlock              & $k' \times (h/s) \times (w/s)$ \\ \bottomrule
\multicolumn{3}{l}{\textbf{Note}: Inverted Residual Attention Block transforming from $k$ to $k'$ chann-} \\
\multicolumn{3}{l}{els, with stride $s$, and expansion factor $t$. dwise means depthwise separable} \\
\multicolumn{3}{l}{convolution.}\\
\vspace{-6 mm}
\end{tabularx}
\vspace{-4mm}
\end{table}

\subsection{MobilePosenet}
\label{subsec:MPPS}

MobilePosenet is an end-to-end model for permanent magnet pose estimation that maps the sensor array-related inputs directly to the position and orientation of the permanent magnet. This approach provides an efficient solution for real-time magnet localization.
Fig. \ref{fig_sim} shows the workflow and network structure of MobilePosenet, with the information flow detailed in Table \ref{tab:IFOM}. MobilePosenet employs a \(6 \times 4 \times 4\) input tensor \((B_{x}, B_{y}, B_{z}, X, Y, Z)\), constructed using theoretical sensor readings from an array of 16 triaxial magnetometers, along with their corresponding sensor positions, to predict the pose of a permanent magnet. The model output comprises two \(1 \times 3\) tensors: one for the magnet's position \((a, b, c)\) and the other for its orientation \((m, n, p)\). Zero-padding is applied to expand the input tensor to a size of \(6 \times 8 \times 8\), ensuring that the original sensor data remains centered. This operation preserves the integrity of the magnetic field information and prevents edge sensor data loss during convolution, thereby enhancing the localization accuracy of the neural network. The padded input is subsequently processed by multiple IRABs. Depthwise separable convolution-based IRAB, coupled with the small input size, renders MobilePosenet an exceptionally lightweight framework. MobilePosenet contains 568,070 number of parameters (NP) and performs 10,565,104 floating-point operations (FLOPs). These computational costs represent only 20.2\% and 50\% of those reported in \cite{su2023amagposenet}. The lower computational overhead indicates that this network could be deployed for high-frequency localization across a broad range of devices with minimal computational hardware demand.

\begin{table}[t]
\caption{INFORMATION FLOW OF MOBILEPOSENET}
\label{tab:IFOM}
\vspace{-4pt}
\centering
\begin{tabular}{|l|l|l|l|l|l|}
\hline
\textbf{Input}           & \textbf{Operator}           & \textbf{t} & \textbf{c} & \textbf{n} & \textbf{s} \\ \hline
$6 \times 4 \times 4$    & ZeroPad2d(2)                & -          & 6          & -          & -          \\ 
$6 \times 8 \times 8$    & ConvBnRelu                  & -          & 32         & 1          & 1          \\ 
$32 \times 8 \times 8$   & IRAB                        & 1          & 16         & 1          & 1          \\ 
$16 \times 8 \times 8$   & IRAB                        & 6          & 32         & 2          & 1          \\ 
$32 \times 8 \times 8$   & IRAB                        & 6          & 64         & 2          & 2          \\ 
$64 \times 4 \times 4$   & IRAB                        & 6          & 128        & 2          & 1          \\
$128 \times 4 \times 4$  & IRAB                        & 1          & 256        & 1          & 1          \\ 
$256 \times 4 \times 4$  & ConvBNReLU                  & -          & 512        & 1          & 1          \\ 
$512 \times 4 \times 4$  & AdaptiveAvgPool2d(1,1)      & -          & 512        & -          & -          \\ 
$512 \times 1 \times 1$  & flatten                     & -          & 512        & -          & -          \\ 
$512 \times 1$           & Linear                      & -          & 3          & -          & -          \\ 
$512 \times 1$           & Linear                      & -          & 3          & -          & -          \\ \hline
\end{tabular}
\par
\vspace{4pt}
\parbox{\columnwidth}{
\textnormal{\textbf{Note}: ConvBnRelu refers to a composite operation consisting of a Conv2d layer, batch normalization, and the ReLU activation function. IRAB denotes an inverted residual attention block. "–" indicates that the layer has no associated parameters. Each row in the table describes a sequence of one or more identical layers, repeated \(n\) times. All layers within the same sequence have the same number \(c\) of output channels. The first layer of each sequence has a stride \(s\), while all subsequent layers use a stride of 1. All spatial convolutions employ \(3 \times 3\) kernels. The expansion factor \(t\) is consistently applied to the input size, as shown in Table \ref{tab:IRAB}.}}
\vspace{-4mm}
\end{table}

\section{EXPERIMENTS}
\label{sec:EXPERIMENTS}
\subsection{Hardware Setup}
\label{subsec:HS}
\begin{figure}[t]
\centering
\includegraphics[width=\columnwidth]{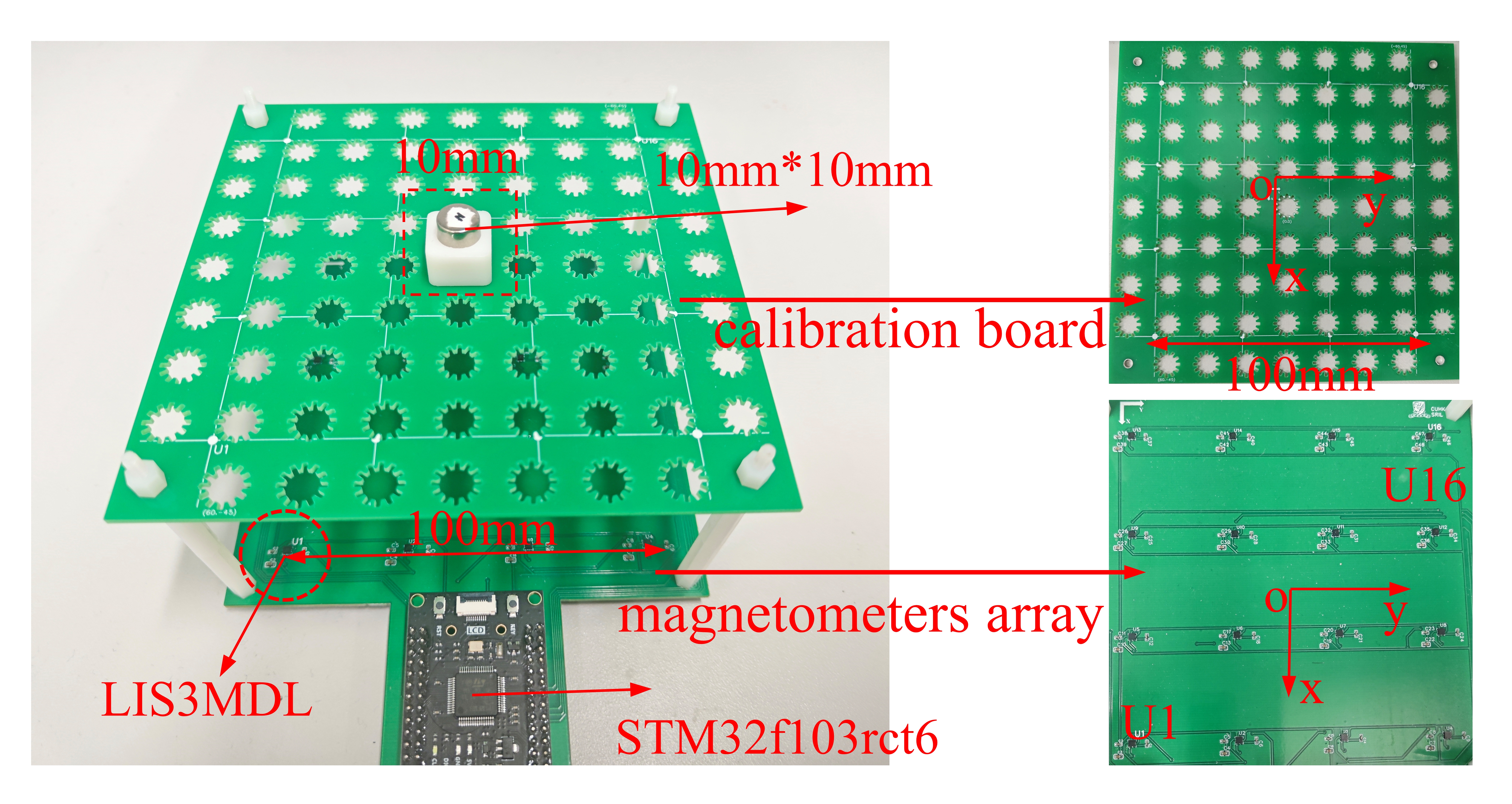}
\vspace{-8mm}
\caption{The experimental platform comprises a triaxial magnetometer array, a calibration board, a plastic magnet shell with a height of 10 mm and an inner diameter of 10 mm, and a cylindrical N35 permanent magnet with a diameter and height of 10 mm (\(B_T = 8.18 \times 10^{-2}\)). The calibration board measures 100 \(\times\) 100 mm, with 15 mm between adjacent holes. The magnetometer array consists of 16 triaxial magnetometers arranged in a \(4 \times 4\) configuration.}
\label{fig_6}
\vspace{-6mm}
\end{figure}

As shown in Fig. \ref{fig_6}, a \(4 \times 4\) magnetometer array was employed to measure the magnetic induction intensities. This array consists of 16 triaxial magnetometers (STMicroelectronics LIS3MDL), evenly soldered onto a printed circuit board (PCB) with \(\frac{100}{3}\) mm spacing. A calibration board measures \(100 \times 100\) mm, with a 15 mm distance between adjacent holes, providing 49 positions for placing the permanent magnets. The permanent magnets have a diameter and height of 10 mm each, and the magnetic flux density \(B_T\) is \(8.18 \times 10^{-2}\) T, calculated following the method described in \cite{su2017investigation}. The distance between the magnetometer array and the calibration board can be adjusted using plastic columns. The magnetometers are configured with a measurement range of ± 1600 \(\mu T\). Data from the magnetometers were collected by a microcontroller unit (STMicroelectronics STM32F103RCT6) and transmitted to a computer running Ubuntu 22.04, equipped with an Intel i7-13700k CPU and an NVIDIA GeForce RTX 4090 GPU.

\subsection{Dataset and Data Preprocessing}
\label{subsec:DDP}
To illustrate the dataset collection process, we first define the workspace in which the dataset resides. The center of the sensor array is defined as the origin of the workspace, with coordinates \((0, 0, 0)\). The sensor plane is defined as the origin plane of the \(z\)-axis in this workspace. The variables \(x\), \(y\), and \(z\) represent the distances of the center point of the permanent magnet from the origin along the respective axes. The rotation angles of the permanent magnet are defined in section~\ref{subsec:MDM}. Since the magnetic dipole model inherently loses information about rotation around its own axis, the system operates in a 5-DOF workspace. The working volume of the system is defined as \(x \in [-45, 45]\) mm, \(y \in [-45, 45]\) mm, \(z \in [40, 120]\) mm, and \(\varphi, \theta \in [-180, 180]^\circ\). The horizontal (x and y axes) range is determined by the distribution of holes in the calibration board. A recent work \cite{guo2022improved} demonstrates that the optimal height range for localization of a cylindrical magnet with dimensions of 10 mm × 10 mm lies between 48 mm and 118 mm. Therefore, a vertical range of 40 mm to 120 mm was selected. In this workspace, two datasets were collected: a theoretical dataset and a real dataset. The theoretical dataset was generated by magnetic dipole model and used for model training, while the real dataset was collected on our self-made platform and used exclusively as the ground truth to evaluate the model's performance and was not involved in the training process.

\begin{figure}[t]
\centering
\includegraphics[width=\columnwidth]{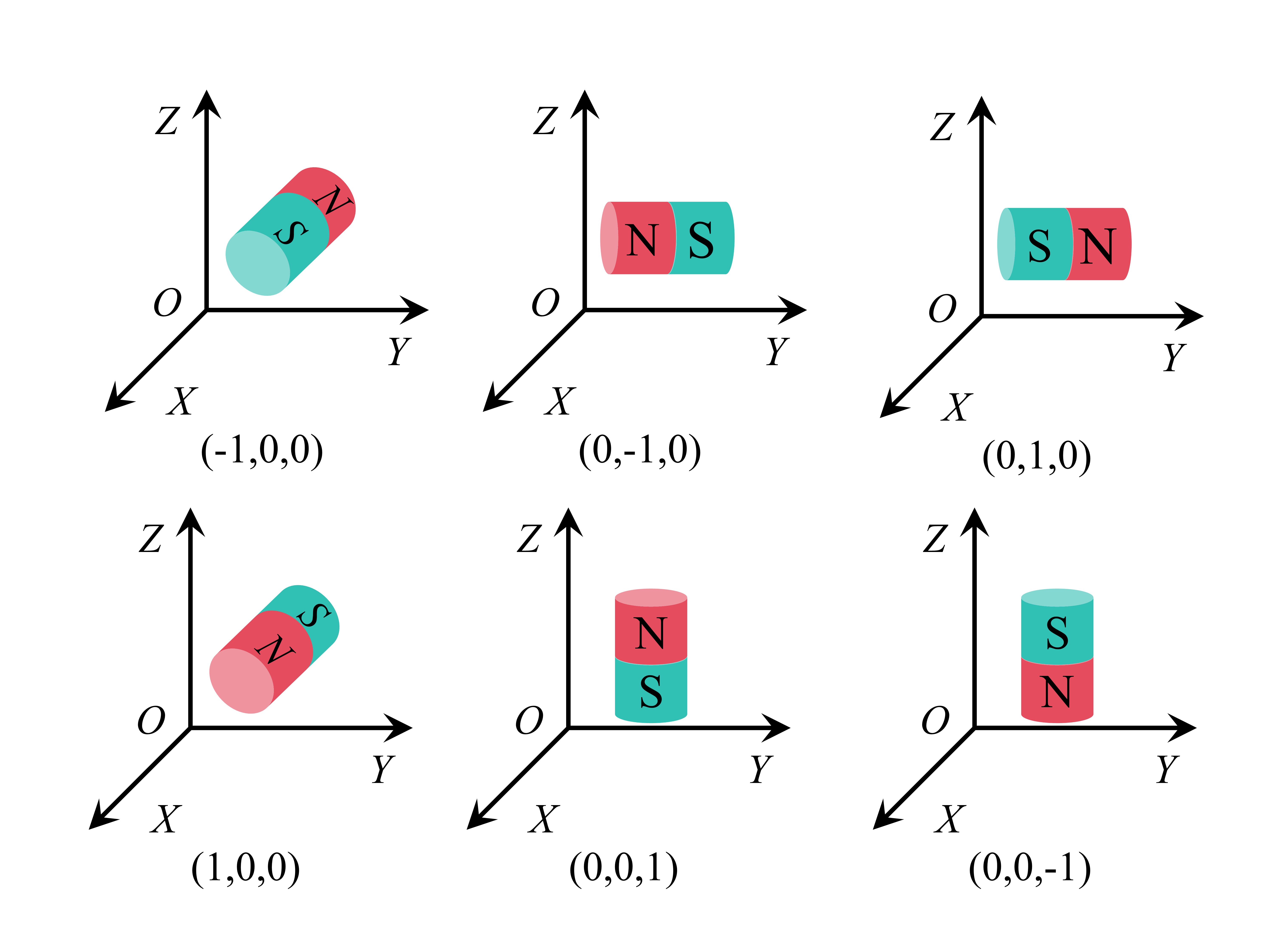}
\vspace{-8mm}
\caption{Six magnet orientations correspond to the boundary values \([-1, 1]\) of the orientation vector \((m, n, p)\).}
\label{fig_7}
\vspace{-6mm}
\end{figure}

The theoretical dataset was calculated based on the magnetic dipole model. Learning-based magnetic tracking heavily depends on the size and resolution of the training dataset \cite{su2023amagposenet}. To ensure that the training space covers the actual working space, the sampling range of the dataset is defined as \(x \in [-50, 50]\) mm, \(y \in [-50, 50]\) mm, and \(z \in [30, 130]\) mm, which slightly exceeds the actual workspace. Sampling points were uniformly distributed across this range, with a spacing of 5 mm. At each location, the rotation angles \(\varphi\) and \(\theta\) were sampled at $\ 30^\circ$ intervals. Additionally, random offsets up to $\pm 5$ mm for position and $\pm 30^\circ$ for angle were applied to each sampling point. This sampling process was repeated three times for each point. Furthermore, to ensure that the boundary values of the orientation vector \((m, n, p)\) within the range \([-1, 1]\) are included in the training dataset, we separately sampled six specific orientations, as shown in Fig. \ref{fig_7}, without adding any offsets. Consequently, the theoretical dataset comprises 4,862,025 data points. From this dataset, 2\% is randomly selected for the test set, another 2\% for the validation set, and the remaining 96\% for the training set.

The real dataset was collected using the calibration board, with the sampling range corresponding to the workspace. The recorded magnetic induction values in the real dataset were the sensor readings minus the geomagnetic field measured when no magnet was nearby. Sampling planes were established along the z-axis from 40 to 120 mm at 10 mm intervals. Each plane contains 49 uniformly distributed sampling points, with 15 mm spacing between adjacent points. The sensor readings were recorded for six magnet poses shown in Fig.~\ref{fig_7} at each point. The real dataset consists of 2,646 pieces of data. 

Notably, there are differences between real data and the corresponding theoretical data due to manufacturing errors, inaccuracies of the magnetic dipole model, and sensor saturation. The standard deviation of this differences is \( \pm 15.27 \, \mu T \), with the maximum deviation of \( 647 \, \mu T \) at the height of 40 mm (primarily due to sensor saturation). 

\subsection{ Network Training Detail and Evaluation Metrics}
\label{subsec:NTDEM}
MobilePosenet was implemented using PyTorch 2.1 and CUDA 11.8. The model was trained using the Adam optimizer over 256 epochs. In the training process, different random noises were added to the input sensor readings at each epoch. The initial learning rate was set to \(1 \times 10^{-4}\), and the cosine annealing schedule was employed to gradually reduce the learning rate to \(0\) over the course of training. The training was conducted on the training set of the theoretical dataset. The model weights corresponding to the best-performing epoch on the validation set were saved as the final model weights. The batch size was set to 3200 and the total training time was approximately 6.4 hours.

To better adjust the loss function and evaluate the localization performance of the model, we separately calculate the position error and orientation error. The position error \(E_p\) and the orientation error \(E_O\) can be expressed as follows. 
\begin{equation}
\label{deqn_ex1a}
{{E}_{\text{p}}}=\sqrt{{{({{a}_{r}}-{{a}_{p}})}^{2}}+{{({{b}_{r}}-{{b}_{p}})}^{2}}+{{({{c}_{r}}-{{c}_{p}})}^{2}}}
\end{equation}
\begin{equation}
\label{deqn_ex1a}
{{E}_{o}}=\sqrt{{{({{m}_{r}}-{{m}_{p}})}^{2}}+{{({{n}_{r}}-{{n}_{p}})}^{2}}+{{({{p}_{r}}-{{p}_{p}})}^{2}}}
\end{equation}
where $(a_r, b_r, c_r,m_r,n_r,p_r)$ is the ground truth and $(a_p, b_p, c_p,m_p,n_p,p_p)$ is the predicted value.
The orientation error can also be expressed as an angle error, calculated using \(2 \sin^{-1}\left(\frac{E_O}{2}\right)\). In the subsequent evaluations, the angle error in this form is reported instead since it is a more commonly used representation.

Considering that the working space size is defined as $90 \times 90 \times 80$ mm, the data distribution of position and orientation is different. To better balance the network's ability to predict the position and orientation of permanent magnets, we add weight $\beta$ to $E_p$ to balance the loss proportion. The final training loss function is shown as follows:
\begin{equation}
\label{deqn_ex1a}
{{E}_{total}}=\beta {{E}_{p}}+{{E}_{O}}
\end{equation}

After model training, we evaluated the model on the test set of the theoretical dataset. The trained model achieved a mean position error of 0.88 mm and an angle error of 0.79°.

\subsection{ Evaluation of Localization Accuracy}
\label{subsec:ELA}
This section evaluates the localization accuracy of MobilePosenet using the real dataset as the ground truth. To demonstrate the effectiveness of the proposed approach, we compared its performance against the LM algorithm by analyzing localization errors on the same datasets. Considering that the accuracy and convergence of the LM algorithm is sensitive to the initial guess, two sets of biases were applied to the ground truth as the initial guesses for the LM algorithm, as detailed below:
\begin{enumerate}
    \item \textbf{bias1}: (3 mm, 3 mm, -3 mm, 0.2, -0.2, 0.2)
    \item \textbf{bias2}: (20 mm, -20 mm, 20 mm, 0.3, 0.3, -0.3)
\end{enumerate}

\begin{figure}[t]
\centering
\includegraphics[width=\columnwidth,keepaspectratio]{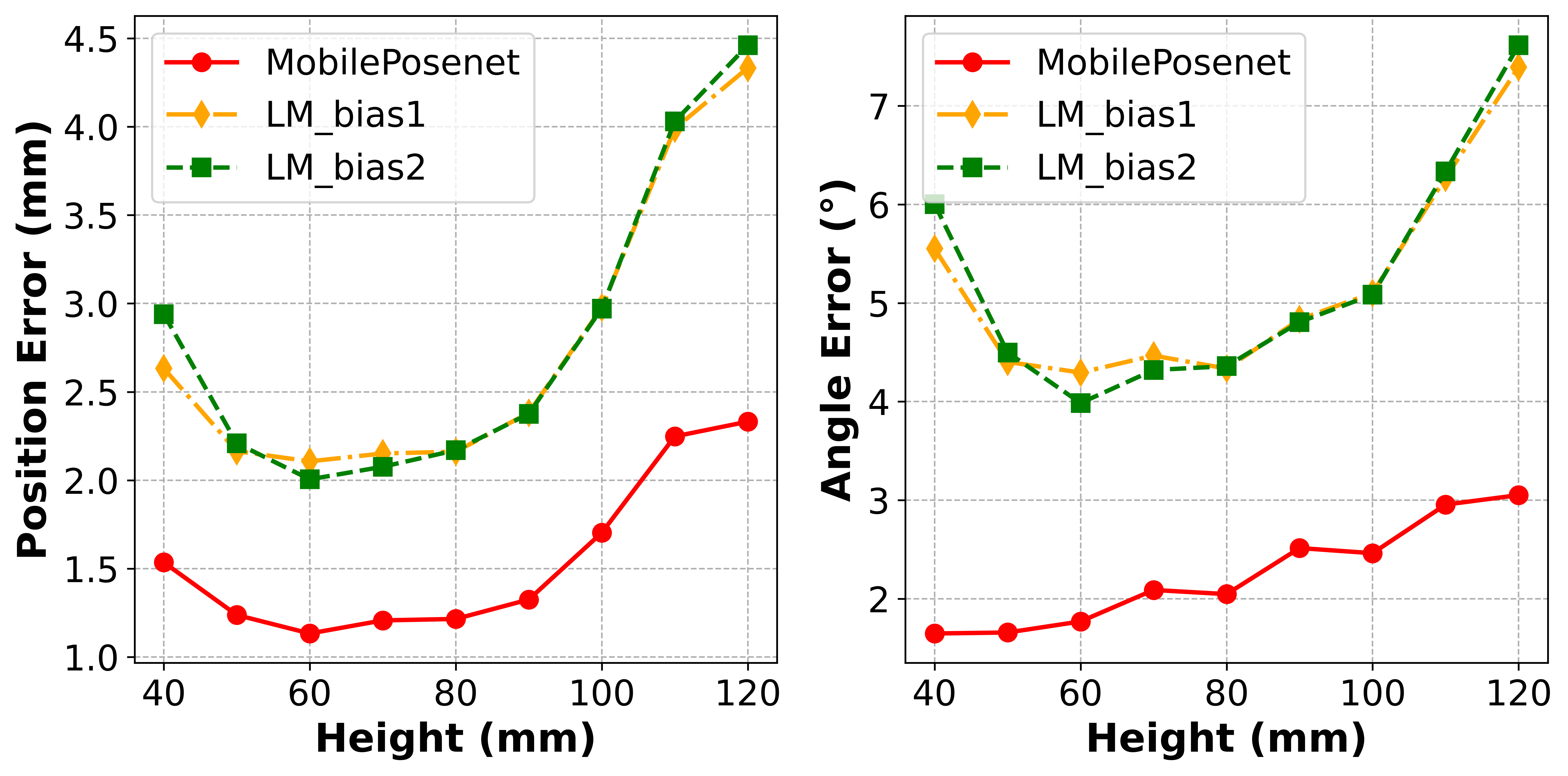}
\vspace{-6mm}
\caption{Positioning errors at various heights. The height refers to the distance between the permanent magnet center and the sensor plane.}
\label{fig_17}
\vspace{-4mm}
\end{figure}

Figure~\ref{fig_17} presents the localization errors of MobilePosenet at various heights in the real dataset. The average position error across all samples is \(1.54 \pm 1.03\) mm, with an angle error of \(2.24 \pm 1.84^\circ\). In contrast, the LM algorithm, with an initial position bias of bias1, exhibits an average position error of \(2.76 \pm 1.58\) mm and an angle error of \(5.18 \pm 2.70^\circ\). When initialized with bias2, the LM algorithm results in an average position error of \(2.88 \pm 1.73\) mm and an angle error of \(5.28 \pm 2.91^\circ\). In addition to larger errors, the LM algorithm also encountered localization failures, which were not observed with MobilePosenet-based localization. These findings clearly demonstrate the superior performance of MobilePosenet, emphasizing its robustness in avoiding localization failures and achieving higher accuracy even when trained exclusively on theoretical datasets.

Notably, the LM-based algorithm exhibited localization failures under conditions of significant initial position bias or substantial input noise. With an initial position bias of bias2, 17.3\% of the 2646 samples in the real dataset failed to localize, with the failure rate rising to 46.9\% at the height of 40 mm in the near field. These results highlight the limitations of the LM algorithm, likely due to significant inaccuracies in the magnetic dipole model when simulating the near field of permanent magnets \cite{6880337}. In the near field, positioning errors are relatively high due to the limited fitting capabilities of the magnetic dipole model and the occurrence of sensor saturation. As the permanent magnet moves away from the sensor array, the saturation effect diminishes, thereby reducing the simulation error of the magnetic dipole model and consequently decreasing the positioning error. However, as the permanent magnet continues to move farther from the sensor array, the positioning error increases again due to the limitations of the sensors' effective working range and measurement capabilities.

\subsection{ Evaluation of Localization Latency}
\label{subsec:ELL}
The execution speed of the positioning algorithm is a critical metric in real-world applications. To evaluate the positioning latency of MobilePosenet, we selected a test sample at each height and conducted positioning tests, comparing the latency with that of the LM algorithm at the corresponding point. Since the positioning latency of the LM algorithm is highly dependent on the initial guess, we introduced two sets of biases as initial guesses for the LM algorithm and recorded the corresponding latency for each. The biases used are the same as described in section \ref{subsec:ELA}.

\begin{table}[t]
\caption{Computation Latency of Different Methods}
\vspace{-4pt}
\label{tab:computation_latency}
\centering
\begin{tabular}{>{\raggedright\arraybackslash}m{1.0cm} >{\centering\arraybackslash}m{1.5cm} >{\centering\arraybackslash}m{1.5cm} >{\centering\arraybackslash}m{1.5cm} >{\centering\arraybackslash}m{1.5cm}}
\toprule
Height  & MobilePosenet \newline (CPU) & MobilePosenet \newline (GPU) & LM-bias1 & LM-bias2 \\
\midrule
40 mm   & \multirow{9}{*}{1.85 ms} & \multirow{9}{*}{0.91 ms} & 340 ms & 350 ms \\
50 mm   &                           &                           & 280 ms & -      \\
60 mm   &                           &                           & 190 ms & -      \\
70 mm   &                           &                           & 170 ms & 170 ms \\
80 mm   &                           &                           & 260 ms & 250 ms \\
90 mm   &                           &                           & 160 ms & 190 ms \\
100 mm  &                           &                           & 150 ms & 250 ms \\
110 mm  &                           &                           & 160 ms & 280 ms \\
120 mm  &                           &                           & 150 ms & 300 ms \\
\midrule
Mean    & \textbf{1.85 ms}          & \textbf{0.91 ms}          & 210 ms & 250 ms \\
\bottomrule
\end{tabular}
\begin{tablenotes}
\centering
\footnotesize
\item \textbf{Note:} (-) indicates that the positioning of the permanent magnet failed.
\end{tablenotes}
\vspace{-6mm}
\end{table}

Table \ref{tab:computation_latency} shows that the computational latency of MobilePosenet is significantly lower than that of the LM algorithm. This is because MobilePosenet is a lightweight network based on depth separable convolution, allowing it to run efficiently on various devices without requiring an initial guess or iterative computation. In contrast, the LM algorithm is an iteration-based optimization method, and its computational delay depends heavily on both the initial guess and the noise level in the input data. When there is a substantial deviation in the initial guess or a high noise level, the execution time of the LM algorithm can increase significantly, and it may even fail to converge. In this experiment, the computational latency of the LM algorithm is higher than that reported in other studies, such as in \cite{su2023amagposenet,guo2022improved}. This discrepancy may arise because the convergence rate of this algorithm is sensitive to noise levels. In this study, the input data were not calibrated, resulting in a higher noise level, which consequently increased the computation delay.

\subsection{Comparison With Existing Approaches}
\label{subsec:CWEA}

Tracking and localizing permanent magnets is a highly specialized task. In addition to the algorithms, tracking performance is influenced by hardware differences across tracking systems, such as the size of the permanent magnets, calibration procedures, and manufacturing tolerances. Therefore, comparing errors and execution times across methods holds limited significance without replicating the hardware systems and datasets. To provide a more comprehensive comparison, we have compiled Table \ref{tab:comparison}, which contrasts the implementation requirements and reported performance of our tracking system with several existing SOTA tracking systems and algorithms. Additionally, we reproduced multiple baseline algorithms to compare their performance with MobilePosenet within our tracking system. To ensure a fair comparison, all model inputs were integrated with coordinate information, and different models were trained under the same conditions.

\begin{table*}[t]
\centering
\begin{threeparttable}
\caption{Comparison of Different Localization Systems}
\vspace{-4pt}
\label{tab:comparison}
\begin{tabular}{@{}lcccccccc@{}}
\toprule
System & \begin{tabular}[c]{@{}c@{}}Rely on \\ Initial Guess\end{tabular} & \begin{tabular}[c]{@{}c@{}}Real Data \\ Training\end{tabular} & \begin{tabular}[c]{@{}c@{}}Need \\ Calibration\end{tabular} & \begin{tabular}[c]{@{}c@{}}Magnet \\ Size (mm)\end{tabular} & \begin{tabular}[c]{@{}c@{}}Number of \\ Magnetometers\end{tabular} & \begin{tabular}[c]{@{}c@{}}Position \\ Error (mm)\end{tabular} & \begin{tabular}[c]{@{}c@{}}Angle \\ Error (°)\end{tabular} & \begin{tabular}[c]{@{}c@{}}Execution \\ Time\end{tabular} \\ \midrule
AmagPosenet \cite{su2023amagposenet} & NO & YES & YES & $5 \times 20$ & 25 & 1.87 & 1.89 & 2.08 ms \\

DNN \cite{sebkhi2019deep}   & NO & YES & YES & $4.8 \times 1.6$ & 24 & 1.4 & - & - \\
PKBPNN \cite{lv2019pkbpnn} & NO & YES & YES & $10 \times 10$ & 9 & 3.48 & 4.31 & - \\
ResNet-LM \cite{guo2022improved}  & NO & YES & YES & $10 \times 10$ & 16 & 0.9 & 1.51 & 79.3 ms \\
Linear Algorithm \cite{hu2007linear}  & NO & NO & YES & - & 5 & 5.6 & 1.7 & - \\
LM \cite{4462308} & YES & NO & YES & $6 \times 12$ & 16 & 3.3 & 3 & 200-300 ms \\
\midrule
\textbf{MobilePosenet (Ours)} & \multirow{7}{*}{NO} & \multirow{7}{*}{NO} & \multirow{7}{*}{NO} & \multirow{7}{*}{$10 \times 10$} & \multirow{7}{*}{16} & \textbf{$\bm{1.54 \pm 1.03}$} & \textbf{$\bm{2.24 \pm 1.84}$}  & \textbf{0.91 ms} \\
AlexNet  & & & & & &$10.97 \pm 3.65$ &$5.41 \pm 2.72$ &2.01 ms \\
VGG11 & & & & & &$37.1 \pm 14.6$ &$2.81 \pm 2.02$ &2.27 ms \\
ResNet & & & & & &$2.16 \pm 1.01$ &$2.81 \pm 1.78$ &3.30 ms \\
MobileNetv2 & & & & & &$2.97 \pm 2.62$ &$3.41 \pm 3.36$ &0.61 ms  \\
EfficientNet & & & & & &$2.12 \pm 1.01$ &$2.59 \pm 1.81$ &1.37 ms \\
\bottomrule
\end{tabular}
\begin{tablenotes}
\centering
\footnotesize
\item (-) indicates that the information is not available. The model architectures of all baseline algorithms were adjusted to accommodate input dimensions specific to the magnetic tracking task (\(6 \times 4 \times 4\)). Every network architecture is configured to downsample the input at most once.
\end{tablenotes}
\end{threeparttable}
\vspace{-6mm}
\end{table*}

Table \ref{tab:comparison} summarizes the implementation requirements and reported performance of various tracking systems. Traditional methods, such as the LM algorithm, rely on accurate initial point estimation and low-noise data, often requiring extensive hardware calibration to achieve high-precision positioning. Similarly, existing neural network-based approaches typically depend on real-world datasets for training or fine-tuning, which can be time-consuming and highly dependent on system-specific calibration, potentially limiting their scalability in large-scale applications. In contrast, permanent magnet tracking using MobilePosenet overcomes these limitations by enabling high-precision and high-frequency positioning in an end-to-end manner, without the need for hardware calibration or real data training. The simple implementation conditions of MobilePosenet allow for the generation of diverse training datasets within minutes, thereby making it adaptable to various positioning spaces and systems.

Even without considering its flexible and simple implementation, MobilePosenet demonstrates significant advantages in both accuracy and speed. It achieves a position error of \(1.54 \, \text{mm}\) and an angle error of \(2.24^\circ\), with an execution time of only \(0.91 \, \text{ms}\), making it the fastest among all current SOTA methods. In terms of positioning accuracy, while ResNet-LM slightly outperforms MobilePosenet in both position error (\(0.9 \, \text{mm}\)) and angle error (\(1.51^\circ\)), its execution time is significantly longer (\(79.3 \, \text{ms}\)). It is also limited to predicting only six specific poses of the magnet due to the dataset constraints. Similarly, DNN achieves a position error of \(1.4 \, \text{mm}\), but it is restricted to only 3-DoF position estimation. Compared to AMagposenet, MobilePosenet offers 5-DoF tracking with a lower position error and execution time. Although Amagposenet shows a slight improvement in angle error, this advantage is primarily due to fine-tuning on real-world data. However, the inherent noise randomness in such data increases the risk of overfitting, which can degrade localization accuracy in other regions of the workspace \cite{10599224}. Unlike traditional methods, such as linear algorithms and the LM algorithm, MobilePosenet does not require an initial guess or calibration, yet still delivers superior accuracy, making it a more reliable and efficient solution for permanent magnet tracking applications.

For a fair comparison independent of hardware systems, MobilePosenet was evaluated against baseline algorithms selected from commonly used SOTA convolutional neural network models. The selection process was guided by the fact that our input size, \(6 \times 4 \times 4\), differs significantly from the standard input size of \(3 \times 244 \times 244\), making more complex networks inefficient for processing such small inputs in the magnetic tracking task. Among the selected baseline methods, MobilePosenet achieves the smallest position and angle errors. Although EfficientNet employs a similar channel attention mechanism and utilizes neural architecture search (NAS), its position error (\(2.12 \, \text{mm}\)) and angle error (\(2.59^\circ\)) are higher. While MobileNetV2 offers slightly faster execution times (\(0.61 \, \text{ms}\)), MobilePosenet provides superior accuracy, making it the most effective solution for real-time localization tasks.

Additionally, by reproducing and evaluating various baseline algorithms within our system, valuable insights were obtained about the performance of neural networks in tracking permanent magnets. Our experiments reveal that increasing the number of parameters does not necessarily result in improved localization accuracy, as evidenced by the lower performance of VGG11 compared to AlexNet. Notably, the inclusion of residual connections, as implemented in ResNet, significantly enhances localization accuracy, outperforming conventional convolutional neural networks that lack such connections. While replacing standard convolutions with depthwise separable convolutions reduces both computational costs and execution time, it also leads to a noticeable decline in tracking accuracy. For instance, ResNet achieves superior localization performance compared to MobileNetV2, despite having an execution time of approximately five times longer. Furthermore, the incorporation of attention mechanisms substantially improves localization accuracy, with only a marginal increase in delay. Although EfficientNet shares architectural similarities with MobilePosenet, such as inverted residual block and SEBlock, its slight decrease in accuracy suggests that the inclusion of dropout layers may not contribute to better generalization performance.

\subsection{ Ablation study }
\label{subsec:AS}
To assess the contribution of each component in our model, we conducted a series of ablation experiments. Additionally, we explored the application of the attention mechanism by introducing the Convolutional Block Attention Module (CBAM) \cite{woo2018cbam} to replace the original channel attention mechanism, SEBLock. This modification is intended to determine whether the inclusion of a spatial attention mechanism could more effectively capture the correlations between different sensors, thereby further enhancing the model's localization capabilities. All modifications were trained under identical conditions to ensure a fair comparison.

\begin{table}[t]
\centering
\caption{Ablation Study on Model Performance}
\vspace{-4pt}
\label{tab:ablation_study}
\begin{tabularx}{\columnwidth}{>{\centering\arraybackslash}X >{\centering\arraybackslash}X >{\centering\arraybackslash}X}
\toprule
\textbf{Experiment} & \textbf{Position Error (mm)} & \textbf{Angle Error ($^\circ$)} \\
\midrule
Baseline & $1.54 \pm 1.03$ & $2.24 \pm 1.84$ \\
w/o New Input & $1.94 \pm 1.52$ & $2.69 \pm 2.37$ \\
w/o SEBlock & $2.97 \pm 2.62$ & $3.41 \pm 3.36$ \\
SEBlock $\rightarrow$ CBAM & $1.52 \pm 1.05$ & $2.32 \pm 1.87$ \\
\bottomrule
\end{tabularx}
\vspace{-6mm}
\end{table}

Table \ref{tab:ablation_study} presents the results of the ablation studies. The baseline model exhibits a position error of $1.54 \pm 1.03$ mm and an angle error of $2.24 \pm 1.84^{\circ}$. When the coordinate information of the sensor array is excluded from the input, the position error increases to $1.94 \pm 1.52$ mm, and the angle error rises to $2.69 \pm 2.37^{\circ}$. These results underscore the significance of coordinate information in enhancing the model's localization capabilities.

The channel attention mechanism plays a critical role in enhancing the neural network's ability to accurately locate permanent magnets. After removing the SEBlock, the positional error increases to $2.97 \pm 2.62$ mm, and the angle error increases to $3.41 \pm 3.36^\circ$. When the SEBlock was replaced with the CBAM module to incorporate spatial attention, the model with the CBAM module achieved a positional error of $1.52 \pm 1.05$ mm and an angle error of $2.32 \pm 1.87^\circ$, with a total training time of approximately 12.2 hours. These results indicate that the addition of spatial attention does not enhance the model's localization accuracy but doubles the training time. Therefore, in the final model, the SEBlock is selected over the CBAM module as the attention mechanism.

\section{CONCLUSION}
\label{sec:conclusion}
In this study, we propose MobilePosenet, a lightweight neural network architecture optimized for real-time permanent magnet localization, with accurate, robust, and efficient localization capabilities. MobilePosenet leverages depthwise separable convolutions and a channel attention mechanism to achieve high localization accuracy with minimal computational cost. To the best of our knowledge, this is the first application of these techniques in the context of permanent magnet tracking, enabling a significant reduction in computational complexity without compromising performance. Additionally, we introduce a novel training strategy based entirely on theoretical datasets generated from the magnetic dipole model, eliminating the need for labor-intensive real-world data collection and sensor calibration, thereby enhancing the system’s practicality and flexibility. The innovative input method, which combines triaxial magnetometer readings with coordinate information and random noise, effectively compensates for discrepancies between idealized models and real magnetic fields, thereby enhancing localization accuracy. However, despite its excellent performance, some limitations persist. The reliance solely on theoretical data may lead to performance degradation under external magnetic interference, which we aim to address in future work by adopting hybrid training methods that integrate real and theoretical data to improve robustness in complex environments. 

\bibliographystyle{IEEEtran}
\bibliography{ref11}

\end{document}